\documentclass{article}

\usepackage{arxiv}

\usepackage[utf8]{inputenc} 
\usepackage[T1]{fontenc}    
\usepackage{hyperref}       
\usepackage{url}            
\usepackage{booktabs}       
\usepackage{amsfonts}       
\usepackage{nicefrac}       
\usepackage{microtype}      
\usepackage{amsmath}
\usepackage{cleveref}       
\usepackage{lipsum}         
\usepackage{graphicx}
\usepackage{natbib}
\usepackage{siunitx}
\usepackage{doi}
\usepackage{subcaption}

\title{Data Augmentation Scheme for Raman Spectra with Highly Correlated Annotations}

\date{}

\newif\ifuniqueAffiliation

\usepackage{authblk}

\author[1]{Christoph Lange}%
\author[2]{Isabel Thiele}
\author[2]{Lara Santolin}
\author[2,3]{Sebastian L. Riedel}
\author[1]{Maxim Borisyak}
\author[1,2]{Peter Neubauer}
\author[1]{M. Nicolas Cruz Bournazou}
\affil[1]{KIWI biolab, Technische Universität Berlin, Ackerstr. 76, Berlin 13355, Germany}
\affil[2]{Technische Universität Berlin, Institute of Biotechnology, Chair of Bioprocess
Engineering, Berlin, Germany}
\affil[3]{Berliner Hochschule für Technik, Department VIII – Mechanical Engineering, Event
Technology and Process Engineering, Laboratory of Environmental and Bioprocess
Engineering, Berlin, Germany}

\renewcommand{\shorttitle}{Data Augmentation Scheme for Raman Spectra with Highly Correlated Annotations}

\hypersetup{
pdftitle={Data Augmentation Scheme for Raman Spectra with Highly Correlated Annotations},
pdfsubject={Raman},
pdfauthor={Christoph Lange, Isabel Thiele, Lara Santolin, Sebastian L. Riedel, Maxim
Borisyak, Peter Neubauer and M. Nicolas Cruz Bournazou},
pdfkeywords={Raman Spectroscopy, Convolutional Neural Network, Data Augmentation},
}

\begin{document}
\maketitle

\begin{abstract}
In biotechnology Raman Spectroscopy is rapidly gaining popularity as a process
analytical technology (PAT) that measures cell densities, substrate- and product
concentrations. As it records vibrational modes of molecules it provides that
information non-invasively in a single spectrum. Typically, partial least squares (PLS)
is the model of choice to infer information about variables of interest from the spectra.
However, biological processes are known for their complexity where convolutional
neural networks (CNN) present a powerful alternative. They can handle non-Gaussian
noise and account for beam misalignment, pixel malfunctions or the presence of
additional substances. However, they require a lot of data during model training, and
they pick up non-linear dependencies in the process variables. In this work, we exploit
the additive nature of spectra in order to generate additional data points from a given
dataset that have statistically independent labels so that a network trained on such data
exhibits low correlations between the model predictions. We show that training a CNN
on these generated data points improves the performance on datasets where the
annotations do not bear the same correlation as the dataset that was used for model
training. This data augmentation technique enables us to reuse spectra as training data
for new contexts that exhibit different correlations. The additional data allows for
building a better and more robust model. This is of interest in scenarios where large
amounts of historical data are available but are currently not used for model training.
We demonstrate the capabilities of the proposed method using synthetic spectra of
\textit{Ralstonia eutropha} batch cultivations to monitor substrate, biomass and
polyhydroxyalkanoate (PHA) biopolymer concentrations during of the experiments.
\end{abstract}

\keywords{Raman Spectroscopy, Convolutional Neural Network and Data Augmentation}

\section{Introduction}
Raman spectroscopy gained popularity in biotechnology as it enables measuring process parameter online in a non-invasive manner. It tracks vibrational modes of molecules that reveal information about the cultivation all in one spectrum. While partial least squares is the model of choice to predict concentrations from spectra, convolutional neural networks are used more often (\cite{qi2023recent}). CNNs are able to handle non-Gaussian noise and account for beam misalignment, pixel malfunctions or the presence of additional substances. Still, due to their immense predictive power, CNNs require large amounts of training data. Hence, data augmentation is common training a CNN (\cite{yun2019cutmix}). It prevents the model from overfitting on characteristics of single observations and promotes learning underlying patterns. This additional data improves the generalization of the neural network and has a regularizing effect on the model. \\
Data obtained from a cultivation process typically has strong dependencies. For example, for a batch cultivation, substrate is inversely related to biomass. Machine Learning models are able to learn these dependencies, which allows them to improve biomass predictions using spectral lines of substrate. While this improvement is desirable for making predictions for similar cultivations, quality quickly degrades when the model is applied to a different process. Continuing the example, a fed-batch cultivation would not have such strong dependencies between biomass and substrate, thus, applying a model trained on a batch cultivation to data from a fed-batch cultivation would, most likely, result in biased predictions. \\
In this paper, we propose a method for "erasing" these dependencies from training data, thus, making the resulting model suitable for a much wider range of processes. We evaluate our approach with multiple synthetic datasets, where the scheme of the synthesis was learned from real Raman spectra recorded during cultivations of \textit{Ralstonia eutropha}, which produced the biodegradable polyhydroxyalkanoate (PHA) copolymer poly(hydroxybutyrate-co-hydroxyhexanoate) [P(HB-co-HHx)], with changing monomer compositions, depending e.g. on the substrates used (\cite{santolin2023tailoring}). As \cite{alcantara2023soft} showed this is a challenging task that CNNs can provide additional benefit for.

\section{Material and Methods}

In this setup m-dimensional Raman spectra  $X \in \mathbb{R}^{n \times m}$ are recorded during fermentation experiments \cite{alcantara2023soft}. To use Raman spectroscopy as a PAT tool we use spectra to predict the concentrations of process parameters $Y \in \mathbb{R}^{n \times k}$  that will be called labels using a CNN. 

\subsection{Data Augmentation Scheme}

Neural networks are almost exclusively trained on mini batches. The following algorithm is applied to each batch, hence $n$  is the batch size. We want to remove the correlations from our annotations $Y$ , so we just sample new uncorrelated annotations  $U \in \mathbb{R}^{n \times k}$ from uniform distributions, 
\begin{equation}
u_{ij} \sim U \left(0 , max \left(\{ Y_{lj} , 1 \leq l \leq n \} \right) \right)
\end{equation}
in the range of the correlated annotations $Y$ .  Now we are looking for coefficients  $\Lambda  \in \mathbb{R}^{n \times n} $that yield the sampled annotations.
\begin{equation}
\Lambda Y = U 
\label{labelcombi}
\end{equation}
We need these coefficients for combining the given spectra $X$ to newly generated spectra $\Lambda X$ that correspond to the uncorrelated annotations $U$. For solving equation (\ref{labelcombi}) we use a singular value decomposition (SVD) of the original annotation matrix $Y$.
\begin{equation}
Y = U \Sigma V^T
\end{equation}
Using a right inverse $\Sigma^R$ of $\Sigma$ we obtain the coefficients $\Lambda$ via:
\begin{equation}
\Lambda = U V \Sigma^R U^T
\end{equation}
The mixing procedure changes the measurement noise in a non-linear fashion. Assuming that measurement noise is homoscedastic and Gaussian with known variance $\sigma_l^2$, the variance of the noise for the i-th synthetic sample in the l-th component is:
\begin{equation}
\mathrm{Var}\left[\sum_{j=1}^N \lambda_{ij}x_{jl} \right] = \sum_{j = 1}^{N} \lambda_{ij} \sigma_l^2 = \left(\sum_{j=1}^N \lambda_{ij} \right) \sigma_l^2
\label{variance}
\end{equation}
Analogous expressions can be derived for different kinds of noise. To match the noise of the generated sample $i$ to the original one, we add artificial noise with variance $ 1 - \sum_{j=1}^N \lambda_{ij} $.  Unfortunately, that is not possible when $\sum_{j=1}^N \lambda_{ij}  > 1$. In this case, we simply reject the sample. In the following we refer to this process as filtering.

\begin{figure}
	\centering
		\includegraphics[scale=0.6]{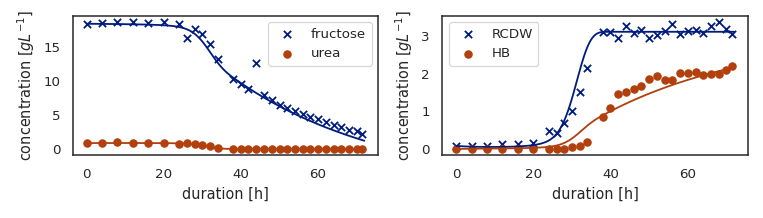}
	\caption{The fit of ODE model to the observations of one cultivation. Left: Substrates. Right products. RCDW = residual cell dry weight, HB = hydroxybutyrate content of the copolymer.}
	\label{fig:ode}
\end{figure}

\subsection{Data Synthesis}
\label{sec:synthesis}

For generating synthetic Raman spectra from \textit{R. eutropha} cultivations, we first use real spectra $X \in \mathbb{R}^{n \times m} $  and \textit{offline} measurements $Y \in \mathbb{R}^{n \times k}$  from two cultivations and decompose them with non-negative matrix factorization (NMF). 
\begin{equation}
X \approx Y H
\end{equation}
Using least squares yields the spectra components $H$ that belong to the respective substance, and we generate new spectra via $c^T H$ with concentrations $c \in \mathbb{R}^k$. \\

The concentrations are obtained from synthetic cultivations that are generated with the help of mechanistic models for \textit{R. eutropha} producing PHA by \cite{khanna2006computer}. We infer parameters $\hat{\theta}$ of the model by  least squares fit to the \textit{offline} measurements from our \textit{R. eutropha} cultivations (figure \ref{fig:ode}). To diversify the set of cultivation parameters we perturb the estimated parameters $\hat{\theta}$ using a gamma distribution. We set $\alpha = \beta = 5$ to ensure an expected value $\mathbb{E} [w_i] = 1$ and preserve similar cultivation dynamics as in the original model.
\begin{equation}
\theta = w^T \hat{\theta}, \quad w_i \sim \Gamma(\alpha, \beta)
\end{equation}
We use the same mechanism to perturb the initial conditions $y_1$.

\begin{figure}
\centering

\begin{minipage}{.55\textwidth}
	  \centering
		\includegraphics[scale=0.35]{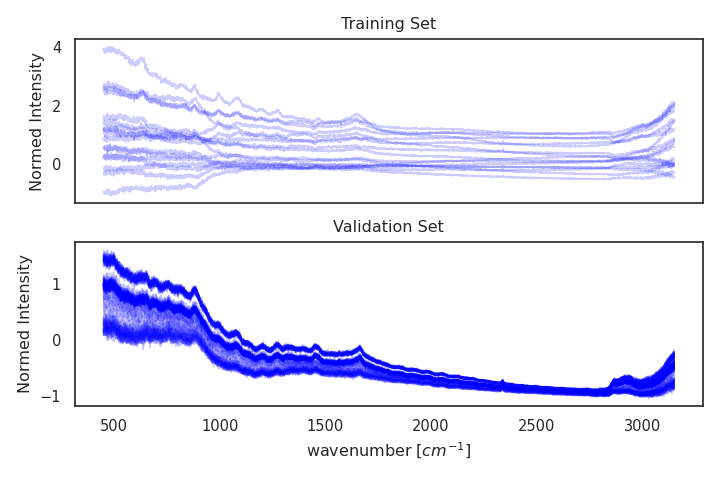}
		\captionsetup{width=0.9\linewidth}
		\captionof{figure}{Normalized spectra generated from the decorrelation algorithm in the training set and unchanged spectra from the validation set.}
	\label{fig:syntheticspectra}
\end{minipage}%
\begin{minipage}{.1 \textwidth}
\end{minipage}
\begin{minipage}{.35\textwidth}
  \centering
		\includegraphics[scale=0.35]{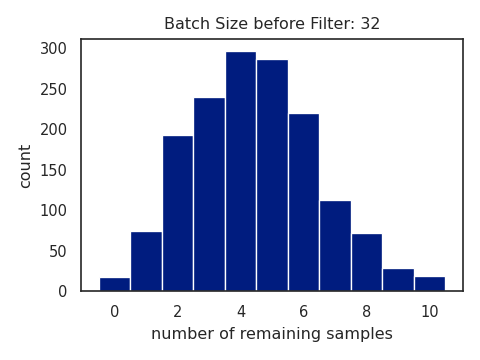}
		\captionsetup{width=0.9\linewidth}
		\captionof{figure}{When filtering out samples with coefficients which norm is greater than 1, we observe this distribution for batch size $32$.}
	\label{fig:filtering}
\end{minipage}
\end{figure}

\subsection{Evaluation Setup}

We use synthetic spectra of \textit{R. eutropha}  batch cultivations that produce the copolymer [P(HB-co-HHx)] with varying monomer composition depending on the available substrates. Canola oil is used for HB and HHx synthesis, whereas fructose only leads to the incorporation of HB monomers into the copolymer (\cite{santolin2023tailoring}). With such a procedure we model realistic changes in statistical dependencies between various substances and measure the performance under novel conditions.

\subsubsection{Datasets}
\label{sec:datasets}
We use the mechanistic model described in \ref{sec:synthesis} to generate cultivations over a period of \SI{72}{\hour}. We use real data from two different cultivations to infer the parameters of the mechanistic model from the \textit{offline} measurements of cell dry weight (CDW), residual cell dry weight (CDW without PHA), fructose, urea, HB and HHx monomer content.  \\
For the training and validation sets we use two different cultivations. Generally, the training set and the validation sets differ in two aspects. On one hand the underlying mechanistic models have different parameters and on the other hand for some validation datasets the fraction of the two carbon substrates differs from the training set ratios according to Table \ref{tab:datasets} which leads to different ratios of HB and HHx. For comparing our algorithm against uncorrelated data, we add the dataset no\_corr, where all concentrations are randomly sampled from a uniform distribution without any structure from a mechanistic model. \\

\begin{table}[h]
	\caption{An overview of the datasets used for the model evaluation. For the training dataset, the percentage is either the first number or the second within one cultivation. The validation datasets are named according to their oil content and “no\_corr” refers to no correlation present. HB, HHx = hydroxybutyrate, hydroxyhexanoate content of the PHA copolymer and all \% refer to wt.\%.}
	\vspace{2mm}
	\begin{tabular}{lrrrrrrrr}
	\hline
	 Dataset & Training   & val 0 & val 2 & val 4 & val 6 & val 8 & val 10 & no\_corr \\
	 \hline
	Canola Oil [\%] & $100 / 0$   & 0 & 20 & 40 & 60 & 80 & 100 & any \\
	Fructose [\%] & $0 / 100 $ & 100 & 80 & 60 & 40 & 20 & 0     & any \\
	Samples & 50,000 & 10,000 & 10,000 & 10,000 & 10,000 & 10,000 & 10,000 & 50,000\\
	HB [\%] &  $ 100 / 80 $ &  100 & 96 & 92 & 88 & 84  & 80 & any \\
	 HHx [\%] & $0 / 20$ & 0 & 4 & 8 & 12 & 16 & 20 & any \\
	\hline
	\end{tabular}
	\label{tab:datasets}
\end{table}

One cultivation and the mechanistic model are depicted in Figure \ref{fig:ode}. All generated cultivations contain all five labels every \SI{3} {\hour} with the corresponding spectra. For the training set we use $2,000$ cultivations and for the validation sets $400$ cultivations respectively. As canola oil is difficult to measure with both Raman spectra and assays, we ignore them for parameter inference and the spectra components.

\begin{figure}
\centering

\begin{minipage}{.55\textwidth}
	  \centering
		\includegraphics[scale=0.35]{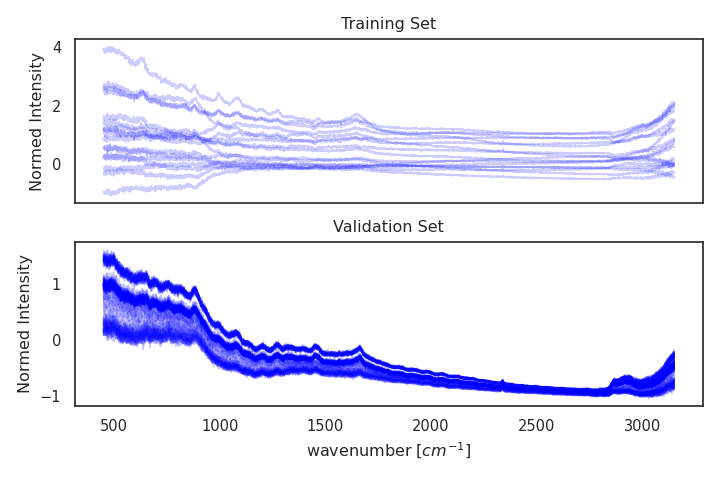}
		\captionsetup{width=0.9\linewidth}
		\captionof{figure}{Normalized spectra generated from the decorrelation algorithm in the training set and unchanged spectra from the validation set.}
	\label{fig:syntheticspectra}
\end{minipage}%
\begin{minipage}{.1 \textwidth}
\end{minipage}
\begin{minipage}{.35\textwidth}
  \centering
		\includegraphics[scale=0.35]{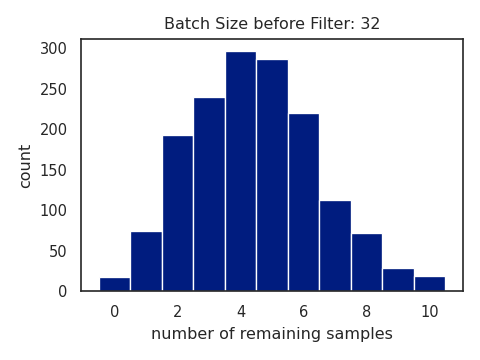}
		\captionsetup{width=0.9\linewidth}
		\captionof{figure}{When filtering out samples with coefficients which norm is greater than 1, we observe this distribution for batch size $32$.}
	\label{fig:filtering}
\end{minipage}
\end{figure}

\subsubsection{Model Architecture}

For all evaluation procedures we use the exact same neural network structure. It is a ReZero architecture by \cite{bachlechner2021rezero} and  \cite{he2016deep} with minor adaptations for one spatial dimension and depthwise separable convolutions from \cite{chollet2017xception} to reduce the number of parameters. The network consists of $8$ residual blocks followed by $3$ fully connected layers with dropout of $0.2$ to make sure the model is properly regularized.

\section{Results}

\subsection{Characteristics of the Decorrelation Algorithm}
Comparing the spectra from the training set obtained by the algorithm to the original spectra from the validation set in figure \ref{fig:syntheticspectra}, we observe that some of the spectra look similar to the ones from the validation set. \\
Some spectra, however, look different, in particular, as if they were inverted. This occurs when some of the mixing coefficient are negative. While such spectra are unrealistic, they do not harm the overall performance, moreover, they might potentially have a regularizing effect on the model. \\
Due to the phenomenon of noise amplification according to equation (\ref{variance})  we filter samples. Looking at figure \ref{fig:filtering} we observe that a high ratio of random samples is filtered out for a batch size of $32$. \\ 
We use six different validation datasets that were generated as described in \ref{sec:datasets}. We train four models in different training setting. For illustration purposes, we include the ideal scenario: one of the models is trained on a dataset with a priori uncorrelated labels (referred to as "no'\_corr"). We trained the others on correlated data for $100$ epochs. \\
According to Table \ref{tab:results} the model trained on the uniform dataset is most successful at transferring its prediction capabilities to different experimental conditions. Among the models trained on the correlated experiments the decorrelation algorithm with filtering performs best. We also highlight the consistency of the proposed method across different validation sets. Not filtering the spectra with excessive noise causes the model to perform even worse than the model only trained on correlated data despite the decorrelation algorithm being in place.

\begin{table}[h]
	\caption{The results of the evaluation procedure. The first three columns describe the training setup of the model. The next six columns depict the mean squared error on the normalized labels of the validation sets described in Table \ref{tab:datasets}. The last column shows the mean of all validation sets.}
	\vspace{2mm}
	\begin{tabular}{lrrrrrrrrr}
	\hline
	 Training Set & Decorrelate  & Filter   & val 0 & val 2 & val 4 & val 6 & val 8 & val 10 & mean \\
	 \hline
	no correlation & no & no    & $0.27$  & $0.18$ & $0.11$ & $0.1$ & $0.05$  & $0.09$  & $0.13$\\
	cultivation like & no & no    & $0.42$  & $0.50$ & $0.41$ & $0.47$ & $0.42$  & $0.48$  & $0.45$\\
	cultivation like & yes & no    & $0.54$  & $0.58$ & $0.52$ & $0.55$ & $0.53$  & $0.58$  & $0.55$\\
	cultivation like & yes & yes    & $0.23$  & $0.25$ & $0.20$ & $0.23$ & $0.20$  & $0.23$  & $0.22$\\
	\hline
	\end{tabular}
	\label{tab:results}

\end{table}

\section{Conclusions}

We propose a data augmentation procedure that allows training robust Machine Learning models on Raman spectra. We show that the procedure "erases" unwanted dependencies in training data, and removes the corresponding biases from the models. The procedure ensures a similar performance of the models across a wide range of cultivation conditions, which dramatically simplified further analysis. \\
We demonstrated performance of our approach on datasets with correlations that differ from the training set. We used the algorithm on Raman spectra of \textit{Ralstonia eutropha}, however, the algorithm exploits only the additive nature of spectral data, and, thus, is agnostic to the spectroscopy methods or the nature of the substances. \\
In practice with the help of our algorithm one can reuse data from old cultivation as training data for a model that infers information from the spectra of a new cultivation setup. This makes models more robust and reduces the number of cultivations needed for new experimental settings.

\section*{Ackknowledgements}
We gratefully acknowledge the financial support of the German Federal Ministry of Education and Research (01DD20002A - KIWI biolab).

\bibliographystyle{unsrtnat}
\bibliography{decorrelation}

\begin{thebibliography}{8}
\providecommand{\natexlab}[1]{#1}
\providecommand{\url}[1]{\texttt{#1}}
\expandafter\ifx\csname urlstyle\endcsname\relax
  \providecommand{\doi}[1]{doi: #1}\else
  \providecommand{\doi}{doi: \begingroup \urlstyle{rm}\Url}\fi

\bibitem[Qi et~al.(2023)Qi, Hu, Jiang, Wu, Zheng, Chen, Liang, Sadi, Zhang, and Chen]{qi2023recent}
Yaping Qi, Dan Hu, Yucheng Jiang, Zhenping Wu, Ming Zheng, Esther~Xinyi Chen, Yong Liang, Mohammad~A Sadi, Kang Zhang, and Yong~P Chen.
\newblock Recent progresses in machine learning assisted raman spectroscopy.
\newblock \emph{Advanced Optical Materials}, page 2203104, 2023.

\bibitem[Yun et~al.(2019)Yun, Han, Oh, Chun, Choe, and Yoo]{yun2019cutmix}
Sangdoo Yun, Dongyoon Han, Seong~Joon Oh, Sanghyuk Chun, Junsuk Choe, and Youngjoon Yoo.
\newblock Cutmix: Regularization strategy to train strong classifiers with localizable features.
\newblock In \emph{Proceedings of the IEEE/CVF international conference on computer vision}, pages 6023--6032, 2019.

\bibitem[Santolin et~al.(2023)Santolin, Thiele, Neubauer, and Riedel]{santolin2023tailoring}
Lara Santolin, Isabel Thiele, Peter Neubauer, and Sebastian~L Riedel.
\newblock Tailoring the hhx monomer content of p (hb-co-hhx) by flexible substrate compositions: scale-up from deep-well-plates to laboratory bioreactor cultivations.
\newblock \emph{Frontiers in Bioengineering and Biotechnology}, 11:\penalty0 1081072, 2023.

\bibitem[Alc{\^a}ntara et~al.(2023)Alc{\^a}ntara, Iannacci, Morbidelli, and Sponchioni]{alcantara2023soft}
Jo{\~a}o Medeiros~Garcia Alc{\^a}ntara, Francesco Iannacci, Massimo Morbidelli, and Mattia Sponchioni.
\newblock Soft sensor based on raman spectroscopy for the in-line monitoring of metabolites and polymer quality in the biomanufacturing of polyhydroxyalkanoates.
\newblock \emph{Journal of Biotechnology}, 377:\penalty0 23--33, 2023.

\bibitem[Khanna and Srivastava(2006)]{khanna2006computer}
Shilpi Khanna and AK~Srivastava.
\newblock Computer simulated fed-batch cultivation for over production of phb: a comparison of simultaneous and alternate feeding of carbon and nitrogen.
\newblock \emph{Biochemical engineering journal}, 27\penalty0 (3):\penalty0 197--203, 2006.

\bibitem[Bachlechner et~al.(2021)Bachlechner, Majumder, Mao, Cottrell, and McAuley]{bachlechner2021rezero}
Thomas Bachlechner, Bodhisattwa~Prasad Majumder, Henry Mao, Gary Cottrell, and Julian McAuley.
\newblock Rezero is all you need: Fast convergence at large depth.
\newblock In \emph{Uncertainty in Artificial Intelligence}, pages 1352--1361. PMLR, 2021.

\bibitem[He et~al.(2016)He, Zhang, Ren, and Sun]{he2016deep}
Kaiming He, Xiangyu Zhang, Shaoqing Ren, and Jian Sun.
\newblock Deep residual learning for image recognition.
\newblock In \emph{Proceedings of the IEEE conference on computer vision and pattern recognition}, pages 770--778, 2016.

\bibitem[Chollet(2017)]{chollet2017xception}
Fran{\c{c}}ois Chollet.
\newblock Xception: Deep learning with depthwise separable convolutions.
\newblock In \emph{Proceedings of the IEEE conference on computer vision and pattern recognition}, pages 1251--1258, 2017.

\end{thebibliography}

\end{document}